\def\year{2020}\relax
\documentclass[letterpaper]{article} 
\usepackage{aaai20}  
\usepackage{times}  
\usepackage{helvet} 
\usepackage{courier}  
\usepackage[hyphens]{url}  
\usepackage{graphicx} 
\usepackage{graphicx}  
\usepackage{amsmath}
\usepackage{algorithm}
\usepackage{amssymb}
\usepackage{subcaption}
\usepackage{multirow}
\usepackage{cancel}

\title{Semantics for Global and Local Interpretation of Deep Neural Networks}

\author{
Jindong Gu \\
The University of Munich\\
Siemens AG, Corporate Technology\\
\texttt{jindong.gu@siemens.com} \\
\And
Volker Tresp \\
The University of Munich\\
Siemens AG, Corporate Technology\\
\texttt{volker.tresp@siemens.com} \\
}

\begin{document}

\maketitle

\begin{abstract}
Deep neural networks (DNNs) with high expressiveness have achieved state-of-the-art performance in many tasks. However, their distributed feature representations are difficult to interpret semantically. In this work, human-interpretable semantic concepts are associated with vectors in feature space. The association process is mathematically formulated as an optimization problem. The semantic vectors obtained from the optimal solution are applied to interpret deep neural networks globally and locally. The global interpretations are useful to understand the knowledge learned by DNNs. The interpretation of local behaviors can help to understand individual decisions made by DNNs better. The empirical experiments demonstrate how to use identified semantics to interpret the existing DNNs.
\end{abstract}

\section{Introduction}
Deep neural networks enable many recent advances in artificial intelligence. Due to the lack of their interpretability, it is difficult to explain their decisions to end users. The neural networks operate on features, which do not correspond to human-interpretable concepts. The neural networks work in a feature space $\boldsymbol{E_f}$ spanned by basis vectors $\boldsymbol{e_f}$ corresponding to neural activations. Humans work in a different vector space $\boldsymbol{E_h}$. The interpretation of the features of neural networks is to map $\boldsymbol{E_f}$ to $\boldsymbol{E_h}$. 

Some existing works associate semantic concepts to individual units \cite{zeiler2014visualizing,zhou2014object}: e.g., in vision one looks for a particular unit $\boldsymbol{e_f^i}$ maximally activated by a specific set of input images. The semantic concept shared by these input images is associated with this particular unit. The specific set of images $\boldsymbol{X}$ satisfy
\begin{equation}
\boldsymbol{X} = arg \max_{x\in \mathbb{I}} \langle \boldsymbol{\phi}(x), \boldsymbol{e_f^i} \rangle
\label{equ:select}
\end{equation}
where $\boldsymbol{X}$ is the set of selected images, $\mathbb{I}$ is a held-out set of unseen images, $\boldsymbol{\phi}(x)$ means a deep feature representation of input image $x$, the one-hot vector $\boldsymbol{e_f^i} \in \mathbb{R}^n$ is a basis vector in the feature space $\boldsymbol{E_f}$, which is associated with the $i$-th hidden unit, and $\max$ in equations \ref{equ:select} and \ref{equ:sevec} means the first N images that maximally activate the $i$-th unit.
We aim to associate semantic concepts with individual vectors in feature space. Formally speaking, given a named semantic concept $c$ (e.g. wheels or cats), a hold-out image dataset $\mathbb{I}$ and a feature extractor $\boldsymbol{\phi}()$ to obtain deep representations, the primary task is to find a vector $\boldsymbol{v}_c = \sum_{i}^{N} b_i \cdot \boldsymbol{e_f^i}$ (linear combination of $\boldsymbol{e_f}$) such that the images $\boldsymbol{X}_c$ that contain the concept $c$ can be selected from $\mathbb{I}$ using
\begin{equation}
\boldsymbol{X}_c  = arg \max_{x\in \mathbb{I}} f(\boldsymbol{\phi}(x), \boldsymbol{v}_c)
\label{equ:sevec}
\end{equation}
where $f()$ is a function measuring the similarity between feature representations and the corresponding semantic vector. The challenge is to find a vector $\boldsymbol{v}_c$ and a meaningful function so that the feature vectors of target images $\boldsymbol{X}_c$ are close to $\boldsymbol{v}_c$. The method to compute the vector $\boldsymbol{v}_c$ and the choice of the function $f()$ will be introduced in Sec. \ref{sec:approach}. The vector $\boldsymbol{v}_c$ associated with a semantic concept is called semantic vector (SeVec) in this paper.

The main contribution of this paper is to obtain semantic vectors by solving an optimization problem. With the obtained SeVecs, we interpret the deep neural networks globally. We quantify the relationship between semantic concepts learned by deep neural networks, e.g., how important the concept $stripe$ is to the concept $zebra$. Furthermore, we explore the multifacetedness of individual semantic concepts. Another contribution is to generate better saliency maps to explain individual decisions made by DNNs using identified semantic vectors.

The next section reviews related work. Sec. \ref{sec:approach} introduces and justifies our method to identify semantic concepts in feature space. Sec. \ref{sec:interpret} interprets the deep convolutional neural network models globally using the obtained SeVecs. Sec. \ref{sec:tdattention} explains individual decisions of DNNs. The last section concludes this paper.

\section{Related Work}
The works \cite{zeiler2014visualizing,simonyan2013deep} confirm the existence of semantic components in Convolutional Neural Networks (CNNs) by illustrating that some filters response to a few images sharing a common concept. \cite{zhou2014object} shows that part of units in the neural network trained for scene classification respond to objects in the scenes. Their work also quantitatively measures the interpretability of deep feature representations by evaluating visualizations. \cite{bau2017network} developed a scalable method to measure the interpretability of deep representations. They measured the alignment between single units and single interpretable concepts without labor-intensive evaluation. In this work, we associate human-interpretable concepts with vectors in feature spaces. 

Many saliency methods have been proposed to explain individual classification decisions by creating saliency maps. The perturbation-based forward propagation approaches perturb individual inputs and observe the impact on later neurons in the network. \cite{zeiler2014visualizing,zintgraf2017visualizing} understand deep features and classifications by analyzing the difference of neuron activations after marginalizing over or perturbing each input patch. The backpropagation-based approaches propagate a relevant signal from a deep layer back into the input space in a single pass, layer-by-layer. The signal thereof can be vanilla gradients \cite{simonyan2013deep}, their variants \cite{springenberg2014striving,zhou2016learning,sundararajan2017axiomatic}, or the combination of gradients and activations \cite{bach2015pixel,shrikumar2017learning}. 

The evaluation of local explanations (saliency maps) has been an active research topic recently \cite{adebayo2018sanity,hooker2018evaluating}. The \textit{Completeness} \cite{bach2015pixel}, \textit{Input Invariance} \cite{kindermans2017reliability}, \textit{Implementation Invariance} \cite{sundararajan2017axiomatic}, \textit{Robustness} \cite{alvarez2018robustness} of saliency methods are explored in literature. Another widely studied property of saliency maps is their class-discriminativity. Concretely, the saliency maps produced by DeconvNet, Gradient Visualization, and Guided Backpropagation are proven to be not class-discriminative \cite{mahendran2016salient}. Given a classification decision, they produce almost the same saliency map for different classes. In this work, we improve the class-discriminativity of explanations using the identified semantic vectors and measure the improvement quantitatively by a generalized Pointing Game.

\section{Semantics in Deep Neural Networks}
\label{sec:approach}
In this section, we describe how to overcome the aforementioned challenge. Given a semantic concept $c$ and a feature extractor $\boldsymbol{\phi}()$, the goal is to compute the direction $\boldsymbol{v}_c$ corresponding the concept $c$ in the feature space. $\boldsymbol{I}_c$ is a set of images containing the concept $c$. The feature extractor $\boldsymbol{\phi}()$ is composed of the first $K$ layers of a deep neural network. The extracted features in the $K$-th layer is $\boldsymbol{\phi}(x)$.

The works \cite{agrawal2014analyzing,dosovitskiy2016inverting} concluded that instead of the precise value, the non-zero patterns of feature representations matter to express the discriminative power and code the semantic meaning. Thus, we search for semantic vectors based on the non-zero patterns of feature representations, i.e., binarized feature representations, which are defined as $\boldsymbol{a_i} =  \boldsymbol{1}_{\boldsymbol{\phi}(x_i)>0}$ where $x_i \in \boldsymbol{I}_c$. In the high-dimensional feature space, the non-zero patterns characterize directions in the space. Thus, the cosine similarity is taken as the function $f()$ to describe the distance between examples in the feature space. The direction of the obtained SeVec $\boldsymbol{v}_c$ should be as close as possible to that of all the binarized feature vectors of $x_i \in \boldsymbol{I}_c$. This requirement is formulated as an optimization problem in Equation \ref{equ:sevec_opt}.
\begin{equation} 
\small
\begin{split}
 \boldsymbol{v}_c  & = arg \max_{|\boldsymbol{v'}| = 1} \sum_{i}^{M} cos\_sim(\boldsymbol{a}_i, \boldsymbol{v'})  \\ 
& = arg \max_{|\boldsymbol{v'}| = 1} (\sum_{i}^{M} \hat{\boldsymbol{a}_i}) * \boldsymbol{v'}  \\
& = arg \max_{|\boldsymbol{v'}| = 1} \boldsymbol{A} * \boldsymbol{v'} \\
\end{split}
\label{equ:sevec_opt}
\end{equation}
where $\hat{\boldsymbol{a_i}} = \frac{\boldsymbol{a_i}}{|\boldsymbol{a_i}|}$,  $\boldsymbol{v'}$ is a linear combination of basis vectors in feature space $\mathbb{R}^n$ and the operation $*$ means dot product. The formula satisfies $\boldsymbol{A} * \boldsymbol{v'}  \leq |\boldsymbol{A}| \cdot \cos(\theta)$ where $\theta$ is the angle between $\boldsymbol{A}$ and $\boldsymbol{v'}$. When $\theta = 0$, i.e., $\boldsymbol{v'} = \frac{\boldsymbol{A}}{|\boldsymbol{A}|}$, the fomular $\boldsymbol{A} * \boldsymbol{v'}$ achieves its maximum. Namely, the optimal solution is $\boldsymbol{v}_c = \frac{\boldsymbol{A}}{|\boldsymbol{A}|}$.

We represent each semantic concept with a single vector in the feature space. However, many semantic concepts are multifaceted. To find the number of facets of a semantic concept $c$, we cluster the feature representations of images $\boldsymbol{I}_c$ using cosine distance-based clustering methods. We found that there is always a dominant cluster containing most of the samples, which indicates that the neural networks map all the images of the concept $c$ into a single direction of feature space and learn invariant representations. This conclusion has been drawn many times in the previous publications \cite{donahue2014decaf,oquab2014learning}. Thus, it is reasonable to represent a semantic concept only using a single vector. The multiple facets of semantic concepts in neural networks will be discussed further in subsection \ref{sec:facet}.

In the optimal solution of equation \ref{equ:sevec_opt}, each element of the computed semantic vector is proportional to the activation rate of the corresponding unit. Namely, the value of an element in a SeVec $\boldsymbol{v}_c$ implies the relevance of the corresponding unit to the concept $c$. A unit with a low activation rate can be highly activated in a particular image, which could be caused by a cluttered background. Although the higher the activation rate of a unit is, the more important it is, a single unit itself with a high activation rate is not enough to represent the concept $c$ (see the experiment in Sec. \ref{subsec:map}).

The computed SeVec $\boldsymbol{v}_c$ corresponds to a single direction in feature space in layer $K$. The vicinity of $\boldsymbol{v}_c$ is defined as $B_r(\boldsymbol{v}_c) = \{\boldsymbol{v} \in \mathbb{R}^n \rvert cos\_dis(\boldsymbol{v}, \boldsymbol{v}_c) < r \}$ using cosine similarity which measures the distance of the two directions in feature space. The feature vector $\boldsymbol{\phi}(x)$ of an image containing the concept $c$ are close to $\boldsymbol{v}_c$. With an identified SeVec $\boldsymbol{v}_c$, the selected images $\boldsymbol{X}_c  = arg \max_{x\in \mathbb{I}} cos\_dis(\boldsymbol{\phi}(x), \boldsymbol{v}_c)$ from a hold-out image set is expected to contain the concept $c$. By using the cosine distance-based nearest neighbor rule, one can partition the feature space into subspaces, each for one semantic concept. Conversely, for a concept $c$, each element of the SeVec $\boldsymbol{v}_c$ specifies the relevance of the corresponding dimension to the concept. 

\begin{figure*}
\begin{minipage}{.5\textwidth}
    \begin{subfigure}[b]{\textwidth}
     \centering
        \includegraphics[scale=0.14]{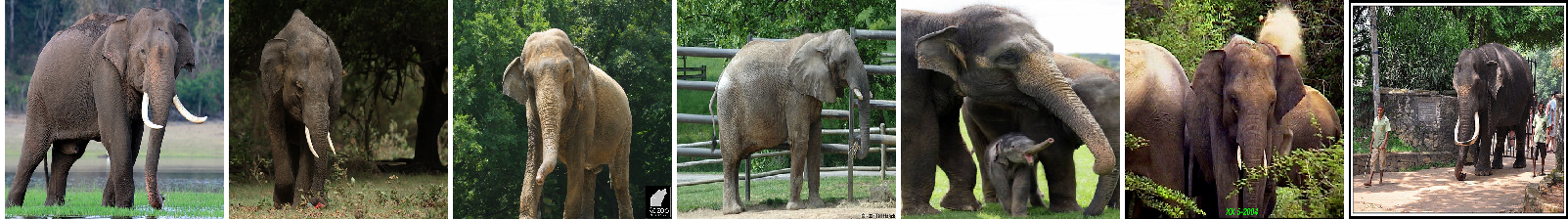}
        \caption{for  the concept \textit{Indian\_elephant}}
    \end{subfigure}
    
    \begin{subfigure}[b]{\textwidth}
    \centering
        \includegraphics[scale=0.14]{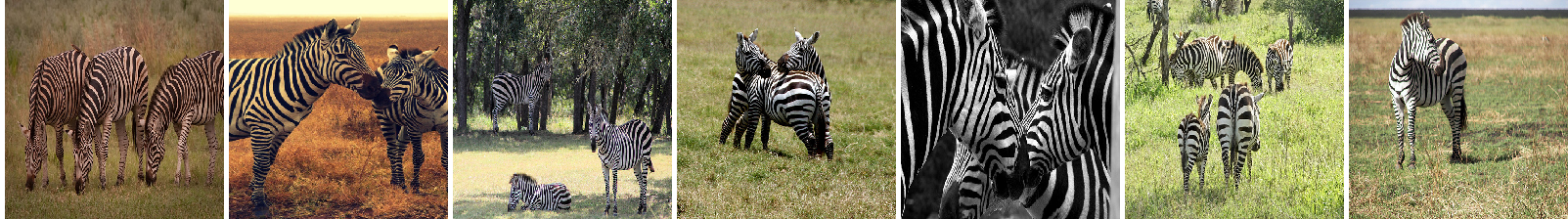}
        \caption{for the concept \textit{Zebra}}
    \end{subfigure}
    
     \begin{subfigure}[b]{\textwidth}
     \centering
        \includegraphics[scale=0.14]{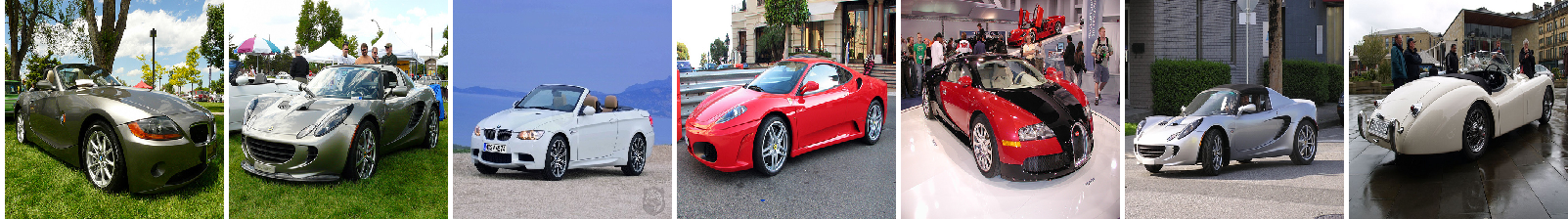}
        \caption{for the concept \textit{Sport\_car}}
    \end{subfigure}
    \caption{For each semantic concept, the images are selected from a hold-out dataset using the obtained SeVecs.}
    \label{fig:retr}
\end{minipage} \hspace{0.2cm}
\begin{minipage}{.5\textwidth}
 \begin{subfigure}[b]{\textwidth}
     \centering
        \includegraphics[scale=0.14]{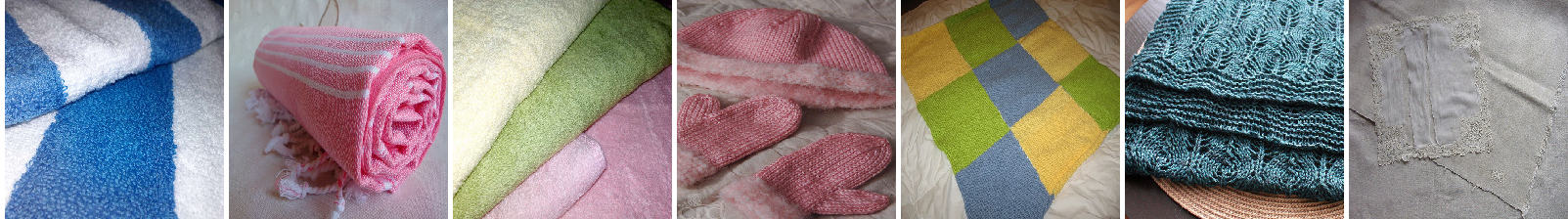}
        \caption{for the material concept \textit{fabric}}
        \label{fig:material_fabric}
    \end{subfigure}
    
    \begin{subfigure}[b]{\textwidth}
    \centering
        \includegraphics[scale=0.14]{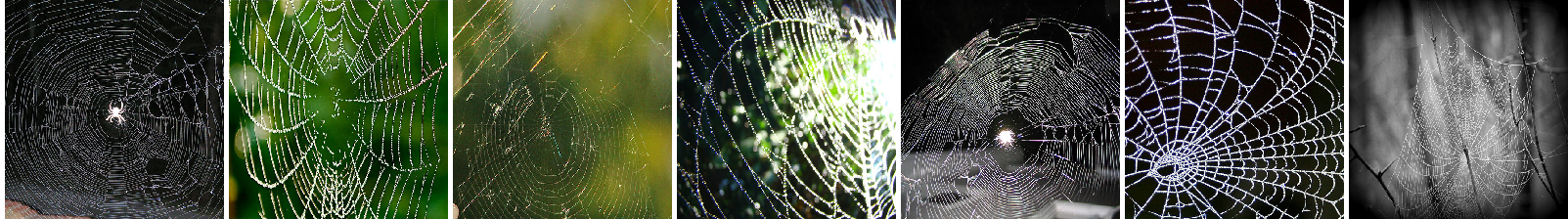}
        \caption{for the textural concept \textit{cobwebbed}}
        \label{fig:texture_cobwebbed}
    \end{subfigure}
    
    \begin{subfigure}[b]{\textwidth}
    \centering
        \includegraphics[scale=0.14]{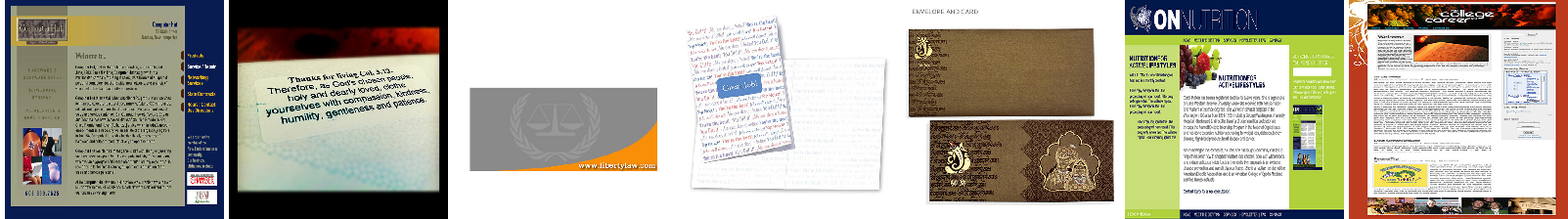}
        \caption{for the color concept \textit{red}}
        \label{fig:color_red}
    \end{subfigure}
    \caption{The selected images using the SeVecs corresponding the low-level visual concepts.}
    \label{fig:retr_others}
    \end{minipage}
\end{figure*}

\subsection{Validation of Semantic Vectors}
In this subsection, we justify the SeVecs resulting from the optimal solution of Equation 3. In the empirical experiments, we used examples of rectifier neural networks, e.g., the VGG16 network \cite{simonyan2014very}. The pre-trained models are taken from the Pytorch framework. Our experiments are conducted in the feature space corresponding to the first fully-connected $(fc1)$ layer of the VGG16 network.

\subsubsection{High-level and Low-level Semantic Concepts} 
The training dataset in ILSVRC 2014 \cite{russakovsky2015imagenet} is labeled with 1000 high-level visual concepts. For each of the 1000 concepts, we compute the corresponding SeVec using the labeled images in the training dataset and select images from a hold-out dataset. For testing, we use the validation dataset as the hold-out dataset. The selected images (top ones) for different concepts are shown in Figure \ref{fig:retr}. The selected images do contain the concepts that SeVecs correspond to. Although unsurprising, the empirical results show that it is meaningful to represent semantic concepts with vectors in the feature space.

We also validate the SeVecs of low-level visual semantic concepts, such as material, texture, and color. To compute the SeVec of a concept, our algorithm requires a number of images containing the concept. However, such labels are not available in the ImageNet database. We turn to other annotated datasets.  For concepts related to texture, material and color, we use the Describable Textures Dataset, Flickr Material Database and Google-512 respectively.

We identify SeVecs of 47 textures, 10 materials and 11 colors using the available labeled images in the three datasets. From the validation dataset of ILSVRC 2014, the images that lie in the vicinity of the corresponding SeVec are selected and shown in Figure \ref{fig:retr_others}. For materials and textures, the selected images contain the corresponding semantic concepts in subfigures \ref{fig:material_fabric} and \ref{fig:texture_cobwebbed}. However, the selected images for color concepts do not show the corresponding color concepts in subfigure \ref{fig:color_red}.

Counter-intuitively, we argue that the color information is not essential in deep representations (e.g., the $fc1$ layer in VGG16). We verify this argument with an ablation study on the validation dataset of ILSVRC 2014. We convert the color images into grey ones and duplicate them in three channels to fit the pre-trained models. The classification performance of VGG16 and AlexNet do not drop significantly. The misclassifications thereof are not necessarily caused by the loss of color information. Compared to original images, the new 'grey' input images are translated in each channel. Such translation can potentially lead to misclassification (due to the vulnerability of neural networks \cite{szegedy2013intriguing,goodfellowexplaining}). The dropped performance is totally recovered after retraining. Hence, we argue that the color information is not essential for the classification in existing deep CNNs. Although VGG16 is not trained with classes in the texture and the material datasets, the obtained SeVecs can still identify the images containing the corresponding concepts.

\begin{table}[t]
      \centering
     \scriptsize
    \begin{tabular}{ | c | c | c |c |}
    \hline
   Associated Neuron & Random Neuron & \textbf{SeVec}  & Permutation of SeVec  \\
     \hline
   1.05e-06 &  -6.72e-07 & \textbf{0.1929} & 4.25e-06\\
     \hline
    \end{tabular}
         \caption{The increased scores of the target output units by modifying the representation using the component associated with semantic concepts (average on 1000 concepts of 50K images).}
     \label{table:drop}
\end{table}

\begin{figure*}[t]
    \centering
    \begin{subfigure}[b]{0.45\textwidth}
    \centering
        \includegraphics[scale=0.4]{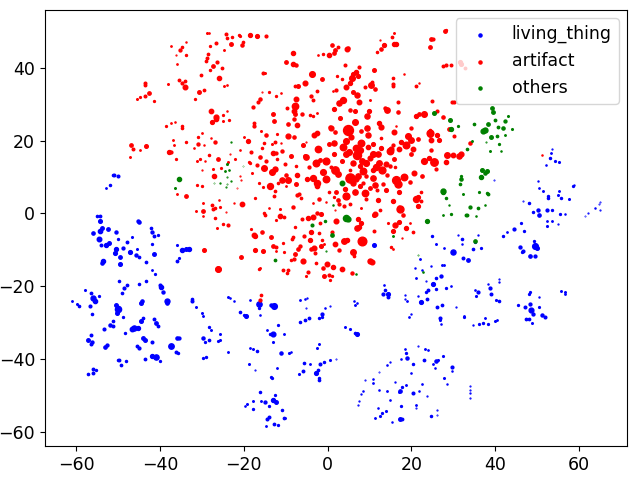}
        \caption{1000 SeVecs of living\_things, artifact and others}
        \label{fig:cluster_all}
    \end{subfigure} \hspace{0.15em}
    \begin{subfigure}[b]{0.45\textwidth}
    \centering
        \includegraphics[scale=0.4]{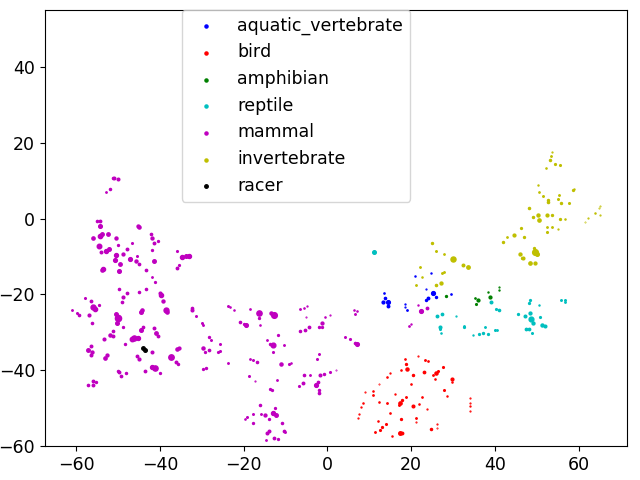}
        \caption{SeVecs belonging to living\_thing (blue points left)}
        \label{fig:subcluster_anoi}
    \end{subfigure}
    \caption{The SeVecs of 1000 semantic concepts (corresponding to 1000 target classes in ImageNet 1k) are visualized using t-SNE \cite{maaten2008visualizing}. Each circle corresponds to a semantic concept, and the size of the circle corresponds to the degree of diversity of the semantic concept.}
    \label{fig:cluster_vectors}
\end{figure*}

\subsubsection{Single Neuron VS. Semantic Vector}
\label{subsec:map}
Does a single neuron or a distributed vector represent the semantic concept in neural networks? In a deep neural network, we observe the changes of the output probabilities of ground-truth classes in case of perturbing the activations of the $K$-th layer. Given an image containing a concept $c$, we identify the single neuron in the $K$-th layer that is most often activated by the concept $c$ in the forward inference, i.e., the maximal element in SeVec $\boldsymbol{v}_c$. We modify the representation by assigning a bigger value (1.5 times the biggest activation in the same layer) to the neuron. If it is the single neuron that corresponds to the semantic concept $c$, the output score of the ground-truth class is supposed to increase. As a comparison, we do the same modification on the activation of a randomly chosen neuron.

We argue that, instead of individual neurons, the directions in feature space correspond to semantic concepts. We modify the representation by setting it closer to the corresponding SeVec (i.e., multiplying $\boldsymbol{1}_{\textbf{v}_c > 0.5}$) and observe the changes of the output of the ground-truth class. For a fair comparison, we also modify the representation by multiplying it with a random permutation of $\boldsymbol{1}_{\textbf{v}_c > 0.5}$.

We conduct experiments on the validation datasets of ILSVRC 2014 containing 50K images. The results of four types of modifications are in Table \ref{table:drop}. The positive values therein are the increased probability value of the corresponding ground-truth class, while the negative values mean the dropped one. If the high activation of the associated neuron means the existence of the visual concept, the output probability of the target should increase. However, the output probabilities of target classes hardly change in case of perturbating a single neuron. The modification using the SeVec increases the confidence by about 20\%. The comparison group has almost no impact on the final output, which ensures that the increased confidence is caused by the semantic meaning instead of the large modification. The results are consistent with the argument that not individual units, but feature vectors are associated with the semantic concepts \cite{szegedy2013intriguing}.

\begin{figure}[h]
    \begin{subfigure}[b]{0.45\textwidth}
    \centering
        \includegraphics[scale=0.14]{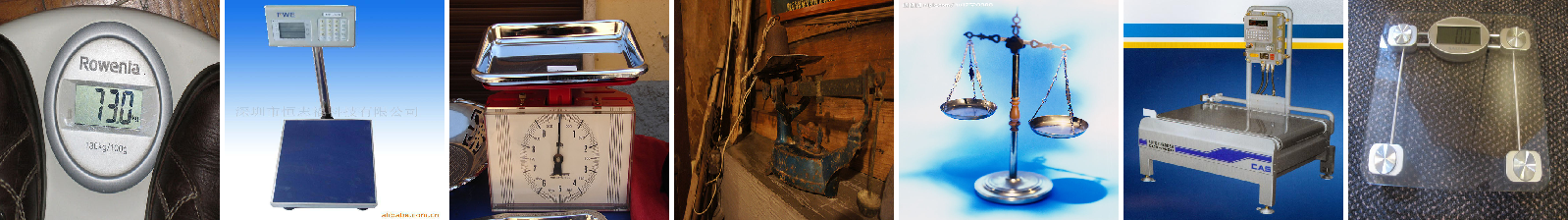}
        \caption{the multi-faceted concept \textit{scale}}
    \end{subfigure}
    \begin{subfigure}[b]{0.45\textwidth}
    \centering
        \includegraphics[scale=0.14]{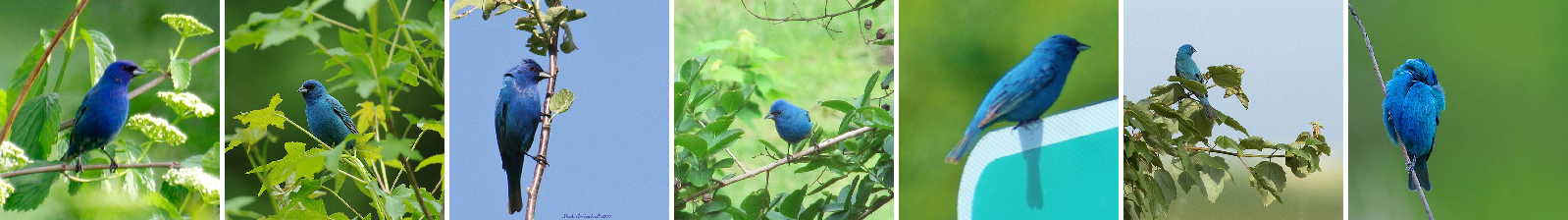}
        \caption{the concept \textit{indigo\_bird} with fewer facets}
    \end{subfigure}
    \caption{The selected images for the concept with the big or small number of facets.}
    \label{fig:multifacet}
\end{figure}

\begin{figure*}[t]
     \centering
        \includegraphics[scale=0.142]{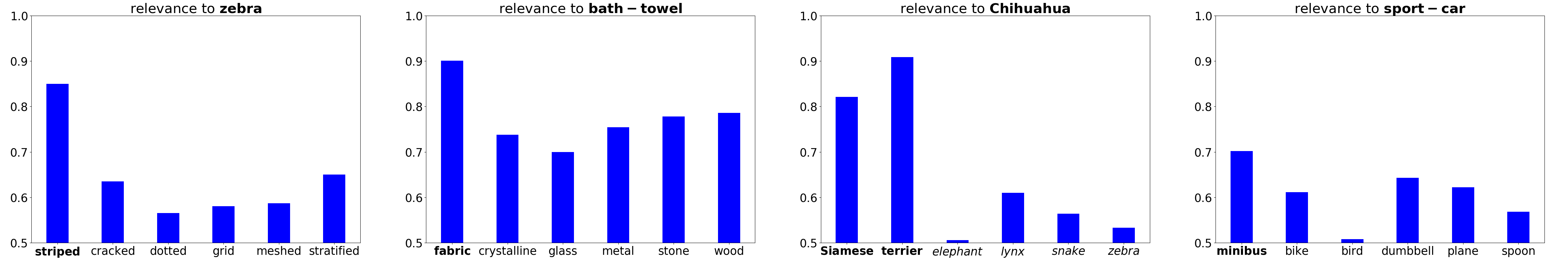}
    \caption{The global interpretation of semantic concepts learned by deep convolutional neural networks. Each figure describes the relationship between semantic concepts. The relevance is calculated on SecVecs using $cosine(\boldsymbol{v}_a, \boldsymbol{v}_b)$.}
    \label{fig:relations}
\end{figure*}

\section{Interpreting Deep CNNs Globally}
\label{sec:interpret}
In this section, we explore global interpretation of neural networks from the perspective of high-level semantic concepts. We answer the following questions: Do CNNs learn semantic-concept hierarchy? How CNNs express the multiple facets of semantic concepts? Do CNNs learn the relationship between high-level concepts and low-level concepts? 

\subsection{Hierarchy of Semantic Concepts in CNNs}
Previous work \cite{donahue2014decaf} visualizes feature vectors of images directly and shows that the feature vectors of the images from similar classes lie near to each other. The semantically similar concepts are visually similar. We aim to verify that semantic concepts learned by CNNs also form a hierarchy. Each of the learned semantic concepts is represented by a SeVec. The distance between their SeVecs characterizes the relationship between two semantic concepts. The SeVecs of similar concepts are expected to be near to each other in the feature space.

In the feature space corresponding the \textit{fc1} layer in VGG16. We analyze the 1000 computed SeVecs together. We project the 1000 SeVecs into two-dimensional space using t-SNE \cite{maaten2008visualizing}. The visualization is shown in Figure \ref{fig:cluster_vectors} where each point corresponds to a SeVec (a concept in ImageNet 1k database). The database is organized hierarchically. Two or more concepts may belong to the same category in a high level of the hierarchy (e.g., \textit{Wader} \& \textit{Turdus} \& $\cdots \rightarrow$ \textit{Bird} $\rightarrow$ \textit{living\_things}).

In the subfigure \ref{fig:cluster_all}, we mark each concept using high-level labels, i.e., \textit{living\_things, artifact and others}. It shows that the SeVecs are well clustered. Each cluster corresponds to a high-level semantic concept. Furthermore, we visualize one of clusters \textit{living\_things} in subfigures \ref{fig:subcluster_anoi} where we mark the semantic concepts with corresponding lower-level labels, i.e., subcategories of \textit{living\_things}. The clear subclusters can also be observed. The hierarchical clusters suggest that VGG16 has learned semantic-concept hierarchy.

\subsection{Multiple Facets of Semantic Concepts in CNNs}
\label{sec:facet}
The activation maximization method \cite{erhan2009visualizing} generates images that maximally activate a neuron without considering its multifacetedness. To understand the multifacetedness of neurons, \cite{nguyen2016multifaceted} separately synthesizes each type of image a neuron fires in response to using prior initialization and regularization methods. The multiple facets of a class are formulated as intra-class knowledge in CNN \cite{wei2015understanding}. Location-variation is expressed in $pool_5$ layer, when content-variation is expressed in $fc_2$ layer in the VGG16 model.

The multiple facets of semantic concepts learned by CNNs are investigated in this work. As discussed before, deep CNNs map most images of a semantic concept into a single direction in the feature space. Clustering in the feature space is not able to define the number of facets of semantic concepts. Thus, instead of the numbers, we start from defining the degree of diversity of semantic concepts. For a semantic concept $c$, the degree of its diversity is defined as $D_c = 1 - \frac{1}{M} \sum_{i}^{M} cos\_sim(\boldsymbol{a}_i, \boldsymbol{v}_c)$ where M is the number of images and $\boldsymbol{a}_i$ is binarized feature representations of an image containing the semantic concept $c$. 

We visualize the degree of diversity of semantic concepts in Figure \ref{fig:cluster_all}. Each circle corresponds to a semantic concept, the size of the circle encodes the $D_c$ of semantic concepts. Different semantic concepts show different degrees of the diversity. The bigger the size is, the larger the number of facets of the corresponding semantic concept is. While some concepts in artifacts show very high diversity, the number of facets of Birds-related concepts is relatively small. More concretely, for instance, the concept $scale$ is multifaceted to a great degree. The degree of diversity of the concepts $indigo\_bird$ is very low. The selected images are shown in Figure \ref{fig:multifacet}.

Does the classification performance of CNNs correlate to the degree of diversity of the concept? We compute the degree of diversity of 1000 semantic concepts $D_c$ and the classification performance of CNNs on the corresponding 1000 classes. The Pearson correlation coefficient and the p-value for testing non-correlation for the two variables, i.e., the $D_c$ and the classification performance, are shown in the table \ref{table:corr}. The table indicates the classification performance of CNNs strongly depends on the degree of multifacetedness of the classes. The higher the degree of the multifacetedness of classes is, the lower the classification performance is.

\begin{table}[h]
      \centering
     \scriptsize
    \begin{tabular}{ | c | c | c |c |}
    \hline
    & & Top 1 accuracy & Top 5 accuracy  \\
     \hline
    \multirow{2}{*}{$D_c$}  &  Correlation Coefficient & -0.5245 & -0.4727\\
     \cline{2-4}
     &  P-value  & 9.395e-72 & 8.485e-57 \\
     \hline
    \end{tabular}
         \caption{The p-value is close to zero, which means the no-correlation assumption is not held. The correlation coefficients with negative values indicate the degree of diversity of semantic concepts is negatively correlated to the top-1 and top-5 classification performance.}
    \label{table:corr}
\end{table}

\subsection{Explanations in CNNs Beyond Saliency Maps}
Most previous methods explain image classification decisions by producing saliency maps. In this experiment, we explain the classification decisions with semantic concepts. The CNNs have learned both high-level and low-level concepts. Do CNNs learn the common sense about the relationship between concepts? The work TCAV \cite{kim2018interpretability} represents semantic concepts using derivatives of the corresponding local linear classifier trained in a specific feature space. The built TCAVs were used to describe the relationship between different concepts. Similarly, we defined the relevance of the concept $a$ to the concept $b$ as $cos\_sim(\boldsymbol{v}_a, \boldsymbol{v}_b)$. We list the relationship between the related concepts in Figure \ref{fig:relations}. The relationship described in the figure corresponds to our common sense. For example, the texture $stripe$ is more important to the concept $zebra$ than other texture concepts.

The individual classification decision can also be explained similarly. The individual explanations with low-level concepts are shown in Table \ref{table:exp_low}. The CNN model predicts the object in the first image as \textit{stone\_wall because it shows cracked texture and its material is stone}. Such explanations with low-level concepts can ensure that the model's predictions base on correct low-level concepts. We also show inappropriate explanations created by this method (marked by a slash). For instance, the \textit{dalmatian} (a species of dog) shows freckled texture, and it is hard to describe what is the material of a \textit{dalmatian}. None of the concepts in Flickr Material Database can be used to describe the material. Similarly, the texture of a whole streetcar is indescribable using the concepts in DTD. Such a simple and novel explanation with low-level concepts can help to understand individual classification decisions.

\begin{table*}[t]
\centering
\footnotesize
{
    \begin{tabular}{ | p{1.5cm} | c | c |  c | c | c | c | c |}
      \hline
     & \includegraphics[width=0.1\textwidth]{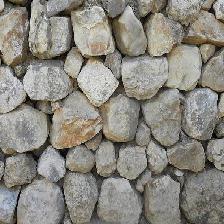} & \includegraphics[width=0.1\textwidth]{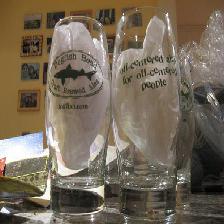} & \includegraphics[width=0.1\textwidth]{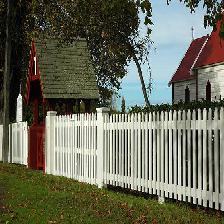} & \includegraphics[width=0.1\textwidth]{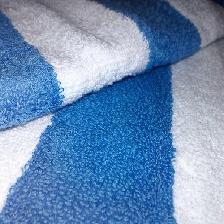} & \includegraphics[width=0.1\textwidth]{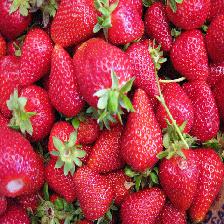} & \includegraphics[width=0.1\textwidth]{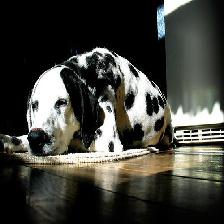} & \includegraphics[width=0.1\textwidth]{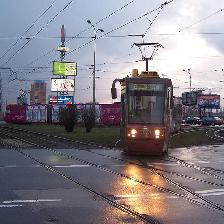} \\
      \hline
      Texcture: &  cracked & crystalline & grooved  & knitted  & bumpy &  freckled & \cancel{porous} \\
      \hline
      Material: & stone & glass & wood  & fabric  & foliage & \cancel{plastic}  & metal \\
      \hline
      Prediction: & \textit{stone\_wall} & \textit{water\_bottle} & \textit{picket\_fence}  & \textit{bath\_towel}  & \textit{strawberry} & \textit{dalmatian} & \textit{streetcar}\\
      \hline
    \end{tabular}
    \caption{The explanations of the classification decisions by illustrating related low-level visual concepts in input images such as the texture, the material of the recognized objects.}
\label{table:exp_low}
}
\end{table*}

\section{Interpreting Deep CNNs Locally}
\label{sec:tdattention}
A large number of saliency methods explain local decisions of deep CNNs via backpropagation processes. They propagate a class-relevant signal back through networks until the input layer and visualizes the signals received by inputs in saliency maps. The existing approaches use no explicit high-level semantic information when propagating the signals. 

After a glance at an image, if we are asked to find the cat in the image, we will search for a cat with a virtual cat pattern in our mind. This phenomenon is explained in the Biased Competition Theory of cognitive science \cite{beck2009top}. In object detection, our visual attention is typically dominated by a goal (high-level semantic information) in a top-down manner. In the feedback loop in brains, the non-relevant neurons are suppressed based on the high-level semantic information.

Inspired by the biased competition theory, we aim to build a similar suppression process using our SeVecs in existing backpropagation-based saliency methods. The obtained SeVecs are applied to suppress the irrelevant internal neurons in backpropagation processes.

Concretely, Guided Backpropagation (GuidedBP) approach combines DeConvNet and Gradient Visualization to produce better saliency maps. The local saliency method describes how the output of the target unit changes for small perturbations around the original input. A global saliency method by multiplying the saliency map with the input describes the marginal effect of a feature on the output with respect to a reference point \cite{ancona2018towards}. In rectifier neural networks, the method Gradient*Input is equivalent to $\epsilon$-LRP \cite{bach2015pixel} and DeepLIFT \cite{shrikumar2017learning}. Without loss of generality, we only consider the local attribution method GuidedBP and the global attribution method Gradient*Input. We will demonstrate our idea on these two simple and representative methods instead of exhaustively generalizing to all existing backpropagation-based saliency methods.

We focus on the \textit{discriminativeness} of explanation maps. For images with multiple objects, a deep CNN have high probabilities for each related class. Given an image $\boldsymbol{x}$ with multiple objects (concepts) $C=\{c_1, c_2, \cdots ,c_n\}$, each of which corresponds to one class in the output layer, the CNN model makes a prediction for the image. The output probability is $O=\{o_1, o_2, \cdots ,o_n\}$. We are supposed to create an explanation map for a unit $o_i$. Explanations should focus on the class-discriminative part of the input image.

Most existing attribution methods suppress the neurons based on thier activations or gradients. The Guided Backpropagation and Gradient*Input are built on the gradient values $\frac{\partial o_i}{\partial \boldsymbol{x}}$. In handling ReLU layers in a backward pass, $R^l = R^{l+1} \textbf{1}_{R^{l+1} > 0\ and\ X^{l} > 0}$ where $R^{n-1} = \frac{\partial o_i}{\partial X^{n-1}}$ and $o_i$ is the $i$-th class-specific unit, the $X^{l}$ are the activations before RuLU layer, and $\textbf{1}$ is the indicator function. However, the neuron activations may caused by cluttered background or other irrelevant concepts. The selection of irrelecant neurons should not depend only on the activations.

We suppress the concept-irrelevant neurons using high-level semantic information. We first compute the SeVec $V_i$ of the class concept $o_i$ in a deep layer. The SeVec we choose is in the feature space of the $pool_5$ in VGG16. The reason for the choice is that the layer maps the spatial information to semantic concepts. This feature space encodes location-variation variance \cite{wei2015understanding}. The SeVec $V_i$ is applied to suppress the irrelevant neurons to filter the irrelevant information in backward pass, $R^l  = R^{l+1} \textbf{1}_{R^{l+1} > 0\ and\ X^{l} > 0\ and\ V_i > 0.5}$.

We only suppress the irrelevant neurons in $pool_5$ layer, which is the most important layer to encode the spatial variation \cite{wei2015understanding}. The feedback CNN \cite{cao2015look} suppresses neurons in all layers by maximizing the output $o_i$. The optimization thereof is an NP-hard problem. The approximated solution proposed in their paper is inefficient, which require many-times backpropagation. Our semantics-based method only requires one backward pass. Our semantic-based method can also be integrated into other saliency methods without much extra cost.

\begin{figure*}
\begin{minipage}{.7\textwidth}
    \centering
    \includegraphics[scale=0.25]{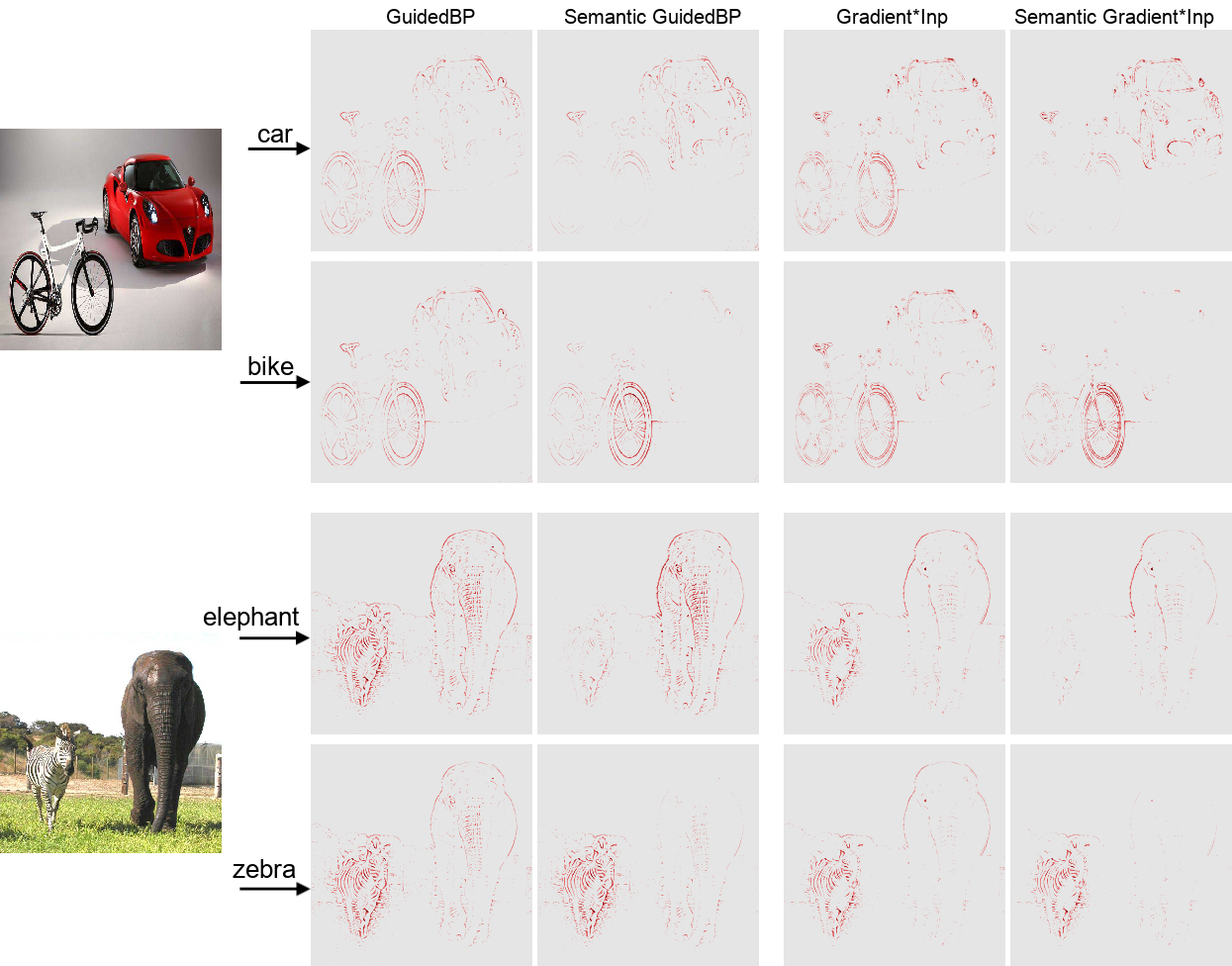}
 \caption{The figure shows explanation maps created by the baseline methods and our semantic versions. Each colum corresponds to a saliency method, and each row shows the saliency maps that surpport classification of a given class.}
    \label{fig:qual_eval}
   \end{minipage} \hspace{0.2cm}
\begin{minipage}{.26\textwidth}
    \centering
    \includegraphics[scale=0.121]{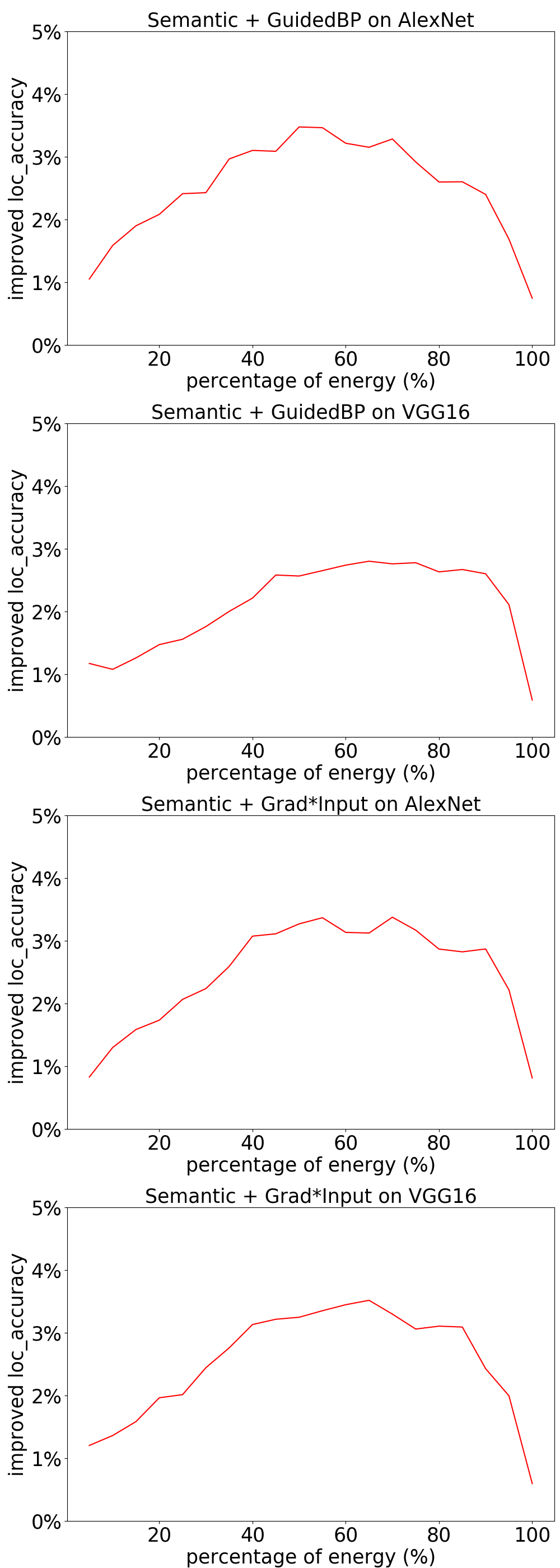}
 \caption{The improved localization accuracy by using our obtained semantic information}
    \label{fig:quan_eval}
  \end{minipage}
\end{figure*}

\subsection{Qualitative Evaluation}
The explanations created by the baseline methods and our semantics-based versions are shown in Figure \ref{fig:qual_eval}. Each column corresponds to a saliency method. For an given image classified by a CNN, the saliency maps are created seperately for two related classes (e.g. $car$ and $bike$ in the first image). The saliency maps are suppose to identify the features supportting a given class. 

For instance, in the first row, the saliency maps should identify the features relevant to $car$. We can observe that the baseline methods (GuidedBP and Gradient*Input) also identify pixels on the $bike$, while our semantic versions (the second and the fourth column) focus more on the pixels on the $car$. Similarly, the second row aims to capture $bike$. While the two baseline approaches visualize both objects and produce visually similar maps for two related classes, our two semantics-based methods are able to identify the relevant object accurately.

\cite{nie2018theoretical} shows that GuidedBP is essentially doing (partial) image recovery which is unrelated to the network decisions. We leverage SeVecs to improve the discriminativeness of GuidedBP. Our semantic GuidedBP does explain the local decision by creating class-discriminative saliency maps. We only compare with baseline approaches to show the improvement since our goal is not to push the state-of-the-art saliency method.

\subsection{Quantitative Evaluation}
The various properties of explanatory saliency maps are explored in publications \cite{adebayo2018sanity,hooker2018evaluating}. In this experiment, we aim to evaluate the discriminativeness of explanatory saliency maps. To quantitatively evaluate the discriminativeness, \cite{zhang2016top} proposes a pointing task where the maximum point of the saliency map is extracted and evaluated. A hit is counted if the maximum point lies in the bounding box of the target object, otherwise a miss is counted. The localization accuracy is measured by $Acc = \frac{\#Hits}{\#Hits+\#Misses}$. We found that the naive pointing at the center of the image shows surprisingly high accuracy. SmoothGrad \cite{smilkov2017smoothgrad} even clips the maximum point to obtain a better visualization. Hence, we generalize the pointing task into a more comprehensive setting. In the new setting, the first step is to preprocess the saliency map by simply thresholding so that $m$ percent energy is kept. A hit is counted if the remaining foreground area (containing relevant pixels) lies in the bounding box of the corresponding target object, otherwise a miss is counted.

The experiments are conducted on two pre-trained deep CNNs (i.e. AlexNet \cite{krizhevsky2012imagenet} and VGG16). Using the baseline approaches and ours, we create saliency maps for ground truth class of each image in the validation dataset of ILSVRC. The improved localization accuracy on the two models is shown in Figure \ref{fig:quan_eval}. When the kept energy is close to 0, the generalized Pointing Game becomes close to the original Pointing Game. In all subplots, the improved localization accuracy is always bigger than zero when the kept energy varies from $0$ to $100\%$. On both models, the localization ability of saliency maps created by our semantic-based method consistently outperforms that of baseline methods. This numerous evidence shows that our SeVecs help to improve the discriminativeness of explanatory saliency maps.

\section{Conclusion}
In this work, we associate human-interpretable semantic concepts with vectors in a feature space, which is formulated as an optimization problem. We apply the semantic vectors obtained from the optimal solution to interpret the convolutional deep neural networks globally and explain individual classification decisions. In addition to understanding positive aspects of them, we also found limitations of the existing deep CNNs. They do not make full use of color information of input images, and perform much worse on the classes with high multifacetedness. Furthermore, a new interpretable semantics-based architecture is desired when we aim to gain trust from users in real-world applications.

\bibliographystyle{aaai20}
\bibliography{aaai20}

\begin{thebibliography}{}

\bibitem[\protect\citeauthoryear{Adebayo \bgroup et al\mbox.\egroup
  }{2018}]{adebayo2018sanity}
Adebayo, J.; Gilmer, J.; Muelly, M.; Goodfellow, I.; Hardt, M.; and Kim, B.
\newblock 2018.
\newblock Sanity checks for saliency maps.
\newblock In {\em NeurIPS},  9505--9515.

\bibitem[\protect\citeauthoryear{Agrawal, Girshick, and
  Malik}{2014}]{agrawal2014analyzing}
Agrawal, P.; Girshick, R.; and Malik, J.
\newblock 2014.
\newblock Analyzing the performance of multilayer neural networks for object
  recognition.
\newblock In {\em ECCV},  329--344.
\newblock Springer.

\bibitem[\protect\citeauthoryear{Alvarez-Melis and
  Jaakkola}{2018}]{alvarez2018robustness}
Alvarez-Melis, D., and Jaakkola, T.~S.
\newblock 2018.
\newblock On the robustness of interpretability methods.
\newblock In {\em 2018 Workshop on Human Interpretability in Machine Learning
  (WHI)}.

\bibitem[\protect\citeauthoryear{Ancona \bgroup et al\mbox.\egroup
  }{2018}]{ancona2018towards}
Ancona, M.; Ceolini, E.; Oztireli, C.; and Gross, M.
\newblock 2018.
\newblock Towards better understanding of gradient-based attribution methods
  for deep neural networks.
\newblock In {\em ICLR}.

\bibitem[\protect\citeauthoryear{Bach \bgroup et al\mbox.\egroup
  }{2015}]{bach2015pixel}
Bach, S.; Binder, A.; Montavon, G.; Klauschen, F.; M{\"u}ller, K.-R.; and
  Samek, W.
\newblock 2015.
\newblock On pixel-wise explanations for non-linear classifier decisions by
  layer-wise relevance propagation.
\newblock {\em PloS one} 10(7):e0130140.

\bibitem[\protect\citeauthoryear{Bau \bgroup et al\mbox.\egroup
  }{2017}]{bau2017network}
Bau, D.; Zhou, B.; Khosla, A.; Oliva, A.; and Torralba, A.
\newblock 2017.
\newblock Network dissection: Quantifying interpretability of deep visual
  representations.
\newblock In {\em CVPR}.

\bibitem[\protect\citeauthoryear{Beck and Kastner}{2009}]{beck2009top}
Beck, D.~M., and Kastner, S.
\newblock 2009.
\newblock Top-down and bottom-up mechanisms in biasing competition in the human
  brain.
\newblock {\em Vision research} 49(10):1154--1165.

\bibitem[\protect\citeauthoryear{Cao \bgroup et al\mbox.\egroup
  }{2015}]{cao2015look}
Cao, C.; Liu, X.; Yang, Y.; Yu, Y.; et~al.
\newblock 2015.
\newblock Look and think twice: Capturing-down visual attention with feedback
  convolutional neural networks.
\newblock In {\em ICCV},  2956--2964.

\bibitem[\protect\citeauthoryear{Donahue \bgroup et al\mbox.\egroup
  }{2014}]{donahue2014decaf}
Donahue, J.; Jia, Y.; Vinyals, O.; Hoffman, J.; Zhang, N.; Tzeng, E.; and
  Darrell, T.
\newblock 2014.
\newblock Decaf: A deep convolutional activation feature for generic visual
  recognition.
\newblock In {\em ICML},  647--655.

\bibitem[\protect\citeauthoryear{Dosovitskiy and
  Brox}{2016}]{dosovitskiy2016inverting}
Dosovitskiy, A., and Brox, T.
\newblock 2016.
\newblock Inverting visual representations with convolutional networks.
\newblock In {\em CVPR},  4829--4837.

\bibitem[\protect\citeauthoryear{Erhan \bgroup et al\mbox.\egroup
  }{2009}]{erhan2009visualizing}
Erhan, D.; Bengio, Y.; Courville, A.; and Vincent, P.
\newblock 2009.
\newblock Visualizing higher-layer features of a deep network.

\bibitem[\protect\citeauthoryear{Goodfellow, Shlens, and
  Szegedy}{2015}]{goodfellowexplaining}
Goodfellow, I.~J.; Shlens, J.; and Szegedy, C.
\newblock 2015.
\newblock Explaining and harnessing adversarial examples.
\newblock In {\em ICLR}.

\bibitem[\protect\citeauthoryear{Hooker \bgroup et al\mbox.\egroup
  }{2018}]{hooker2018evaluating}
Hooker, S.; Erhan, D.; Kindermans, P.-J.; and Kim, B.
\newblock 2018.
\newblock Evaluating feature importance estimates.
\newblock In {\em 2018 Workshop on Human Interpretability in Machine Learning
  (WHI)}.

\bibitem[\protect\citeauthoryear{Kim \bgroup et al\mbox.\egroup
  }{2018}]{kim2018interpretability}
Kim, B.; Wattenberg, M.; Gilmer, J.; Cai, C.; Wexler, J.; Viegas, F.; et~al.
\newblock 2018.
\newblock Interpretability beyond feature attribution: Quantitative testing
  with concept activation vectors (tcav).
\newblock In {\em ICML},  2673--2682.

\bibitem[\protect\citeauthoryear{Kindermans \bgroup et al\mbox.\egroup
  }{2017}]{kindermans2017reliability}
Kindermans, P.-J.; Hooker, S.; Adebayo, J.; Alber, M.; Sch{\"u}tt, K.~T.;
  D{\"a}hne, S.; Erhan, D.; and Kim, B.
\newblock 2017.
\newblock The (un) reliability of saliency methods.

\bibitem[\protect\citeauthoryear{Krizhevsky, Sutskever, and
  Hinton}{2012}]{krizhevsky2012imagenet}
Krizhevsky, A.; Sutskever, I.; and Hinton, G.~E.
\newblock 2012.
\newblock Imagenet classification with deep convolutional neural networks.
\newblock In {\em NeurIPS},  1097--1105.

\bibitem[\protect\citeauthoryear{Maaten and
  Hinton}{2008}]{maaten2008visualizing}
Maaten, L. v.~d., and Hinton, G.
\newblock 2008.
\newblock Visualizing data using t-sne.
\newblock {\em Journal of machine learning research} 9(Nov):2579--2605.

\bibitem[\protect\citeauthoryear{Mahendran and
  Vedaldi}{2016}]{mahendran2016salient}
Mahendran, A., and Vedaldi, A.
\newblock 2016.
\newblock Salient deconvolutional networks.
\newblock In {\em ECCV},  120--135.
\newblock Springer.

\bibitem[\protect\citeauthoryear{Nguyen, Yosinski, and
  Clune}{2016}]{nguyen2016multifaceted}
Nguyen, A.; Yosinski, J.; and Clune, J.
\newblock 2016.
\newblock Multifaceted feature visualization: Uncovering the different types of
  features learned by each neuron in deep neural networks.
\newblock {\em arXiv preprint arXiv:1602.03616}.

\bibitem[\protect\citeauthoryear{Nie, Zhang, and
  Patel}{2018}]{nie2018theoretical}
Nie, W.; Zhang, Y.; and Patel, A.
\newblock 2018.
\newblock A theoretical explanation for perplexing behaviors of
  backpropagation-based visualizations.
\newblock In {\em 2018 Workshop on Human Interpretability in Machine Learning
  (WHI)}.

\bibitem[\protect\citeauthoryear{Oquab \bgroup et al\mbox.\egroup
  }{2014}]{oquab2014learning}
Oquab, M.; Bottou, L.; Laptev, I.; and Sivic, J.
\newblock 2014.
\newblock Learning and transferring mid-level image representations using
  convolutional neural networks.
\newblock In {\em CVPR},  1717--1724.

\bibitem[\protect\citeauthoryear{Russakovsky \bgroup et al\mbox.\egroup
  }{2015}]{russakovsky2015imagenet}
Russakovsky, O.; Deng, J.; Su, H.; Krause, J.; Satheesh, S.; Ma, S.; Huang, Z.;
  Karpathy, A.; Khosla, A.; et~al.
\newblock 2015.
\newblock Imagenet large scale visual recognition challenge.
\newblock {\em ICCV} 115(3):211--252.

\bibitem[\protect\citeauthoryear{Shrikumar, Greenside, and
  Kundaje}{2017}]{shrikumar2017learning}
Shrikumar, A.; Greenside, P.; and Kundaje, A.
\newblock 2017.
\newblock Learning important features through propagating activation
  differences.
\newblock In {\em ICML}.

\bibitem[\protect\citeauthoryear{Simonyan and
  Zisserman}{2015}]{simonyan2014very}
Simonyan, K., and Zisserman, A.
\newblock 2015.
\newblock Very deep convolutional networks for large-scale image recognition.
\newblock In {\em ICLR}.

\bibitem[\protect\citeauthoryear{Simonyan, Vedaldi, and
  Zisserman}{2013}]{simonyan2013deep}
Simonyan, K.; Vedaldi, A.; and Zisserman, A.
\newblock 2013.
\newblock Deep inside convolutional networks: Visualising image classification
  models and saliency maps.
\newblock In {\em ICLR}.

\bibitem[\protect\citeauthoryear{Smilkov \bgroup et al\mbox.\egroup
  }{2017}]{smilkov2017smoothgrad}
Smilkov, D.; Thorat, N.; Kim, B.; Vi{\'e}gas, F.; and Wattenberg, M.
\newblock 2017.
\newblock Smoothgrad: removing noise by adding noise.
\newblock {\em arXiv preprint arXiv:1706.03825}.

\bibitem[\protect\citeauthoryear{Springenberg \bgroup et al\mbox.\egroup
  }{2014}]{springenberg2014striving}
Springenberg, J.~T.; Dosovitskiy, A.; Brox, T.; and Riedmiller, M.~A.
\newblock 2014.
\newblock Striving for simplicity: The all convolutional net.
\newblock In {\em ICLR}.

\bibitem[\protect\citeauthoryear{Sundararajan, Taly, and
  Yan}{2017}]{sundararajan2017axiomatic}
Sundararajan, M.; Taly, A.; and Yan, Q.
\newblock 2017.
\newblock Axiomatic attribution for deep networks.
\newblock In {\em ICML}.

\bibitem[\protect\citeauthoryear{Szegedy \bgroup et al\mbox.\egroup
  }{2014}]{szegedy2013intriguing}
Szegedy, C.; Zaremba, W.; Sutskever, I.; Bruna, J.; Erhan, D.; Goodfellow,
  I.~J.; and Fergus, R.
\newblock 2014.
\newblock Intriguing properties of neural networks.
\newblock In {\em ICLR}.

\bibitem[\protect\citeauthoryear{Wei \bgroup et al\mbox.\egroup
  }{2015}]{wei2015understanding}
Wei, D.; Zhou, B.; Torrabla, A.; and Freeman, W.
\newblock 2015.
\newblock Understanding intra-class knowledge inside cnn.
\newblock {\em arXiv preprint arXiv:1507.02379}.

\bibitem[\protect\citeauthoryear{Zeiler and
  Fergus}{2014}]{zeiler2014visualizing}
Zeiler, M.~D., and Fergus, R.
\newblock 2014.
\newblock Visualizing and understanding convolutional networks.
\newblock In {\em ECCV},  818--833.
\newblock Springer.

\bibitem[\protect\citeauthoryear{Zhang \bgroup et al\mbox.\egroup
  }{2016}]{zhang2016top}
Zhang, J.; Lin, Z.; Brandt, J.; Shen, X.; and Sclaroff, S.
\newblock 2016.
\newblock Top-down neural attention by excitation backprop.
\newblock In {\em ECCV},  543--559.
\newblock Springer.

\bibitem[\protect\citeauthoryear{Zhou \bgroup et al\mbox.\egroup
  }{2014}]{zhou2014object}
Zhou, B.; Khosla, A.; Lapedriza, A.; Oliva, A.; and Torralba, A.
\newblock 2014.
\newblock Object detectors emerge in deep scene cnns.
\newblock {\em arXiv preprint arXiv:1412.6856}.

\bibitem[\protect\citeauthoryear{Zhou \bgroup et al\mbox.\egroup
  }{2016}]{zhou2016learning}
Zhou, B.; Khosla, A.; Lapedriza, A.; Oliva, A.; and Torralba, A.
\newblock 2016.
\newblock Learning deep features for discriminative localization.
\newblock In {\em CVPR},  2921--2929.

\bibitem[\protect\citeauthoryear{Zintgraf \bgroup et al\mbox.\egroup
  }{2017}]{zintgraf2017visualizing}
Zintgraf, L.~M.; Cohen, T.; Adel, T.; and Welling, M.
\newblock 2017.
\newblock Visualizing deep neural network decisions: Prediction difference
  analysis.
\newblock In {\em ICLR}.

\end{thebibliography}

\end{document}